\documentclass[10pt,twocolumn,letterpaper]{article}

\usepackage[pagenumbers]{cvpr} 

\usepackage{graphicx}
\usepackage{amsmath}
\usepackage{amssymb}
\usepackage{booktabs}

\usepackage{times}
\usepackage{epsfig}
\usepackage{graphicx}
\usepackage{amsmath}
\usepackage{amssymb}

\usepackage{comment}
\usepackage{color}

\usepackage{booktabs}
\usepackage{algorithm}
\usepackage{algorithmic}
\usepackage{multirow}
\usepackage{tabularx}
\usepackage{verbatim}
\usepackage{subcaption}
\usepackage{pifont}
\usepackage{soul}

\usepackage[dvipsnames]{xcolor}
\usepackage{colortbl}

\definecolor{lightblue}{rgb}{0.85, 0.95, 1.0}

\usepackage[pagebackref,breaklinks,colorlinks,citecolor=cvprblue]{hyperref}

\usepackage[capitalize]{cleveref}
\crefname{section}{Sec.}{Secs.}
\Crefname{section}{Section}{Sections}
\Crefname{table}{Table}{Tables}
\crefname{table}{Tab.}{Tabs.}
\usepackage[dvipsnames]{xcolor}

\definecolor{cvprblue}{rgb}{0.21,0.49,0.74}
\definecolor{demored}{RGB}{239, 69, 31}
\definecolor{demogreen}{RGB}{112, 173, 71}

\def\paperID{****} 
\def\confName{CVPR}
\def\confYear{2024}

\title{CuMo: Scaling Multimodal LLM with Co-Upcycled Mixture-of-Experts}

\begin{document}

\author{
    Jiachen Li\textsuperscript{1*} \quad
    Xinyao Wang\textsuperscript{2\textdagger} \quad
    Sijie Zhu\textsuperscript{2} \quad
    Chia-Wen Kuo\textsuperscript{2} \quad
    Lu Xu\textsuperscript{2} \quad \\
    Fan Chen\textsuperscript{2} \quad
    Jitesh Jain\textsuperscript{1} \quad
    Humphrey Shi\textsuperscript{1\textdagger} \quad
    Longyin Wen\textsuperscript{2} \\
{\textsuperscript{1}SHI Labs $@$ Georgia Tech \& UIUC,
\textsuperscript{2}ByteDance Inc.}\\
{\small \textbf{\url{https://github.com/SHI-Labs/CuMo}}}
}

\maketitle

\let\thefootnote\relax\footnote{{\textsuperscript{*} Work done during an internship at ByteDance Inc., San Jose, CA. Correspondence to X. Wang (xinyao.wang@bytedance.com) and H. Shi.}}

\begin{abstract}
Recent advancements in Multimodal Large Language Models (LLMs) have focused primarily on scaling by increasing text-image pair data and enhancing LLMs to improve performance on multimodal tasks. However, these scaling approaches are computationally expensive and overlook the significance of efficiently improving model capabilities from the vision side. 
Inspired by the successful applications of Mixture-of-Experts~(MoE) in LLMs, which improves model scalability during training while keeping inference costs similar to those of smaller models, we propose \textbf{CuMo}, which incorporates \textbf{C}o-\textbf{u}pcycled Top-K sparsely-gated \textbf{M}ixture-\textbf{o}f-experts blocks into both the vision encoder and the MLP connector, thereby enhancing the multimodal LLMs with neglectable additional activated parameters during inference.
CuMo first pre-trains the MLP blocks and then initializes each expert in the MoE block from the pre-trained MLP block during the visual instruction tuning stage, with auxiliary losses to ensure a balanced loading of experts.
CuMo outperforms state-of-the-art multimodal LLMs across various VQA and visual-instruction-following benchmarks within each model size group, all while training exclusively on open-sourced datasets.
The code and model weights for CuMo are open-sourced at \href{https://github.com/SHI-Labs/CuMo}{https://github.com/SHI-Labs/CuMo}.

\end{abstract}

\section{Introduction}
The advent of GPT-4V~\cite{gpt4v} has sparked excitement within open-source communities to transform large language models~(LLM) into multimodal LLMs. Recent multimodal LLMs~\cite{dai2023instructblip, liu2023llava, bai2023qwen} typically integrate pre-trained vision encoders and LLMs with visual instruction tuning data to fine-tune the pre-trained LLMs, enhancing their visual understanding capabilities. To further scale up multimodal LLMs, previous efforts~\cite{liu2023improvedllava, liu2024llavanext, li2024minigemini,mckinzie2024mm1, chen2023sharegpt4v, lin2023vila} primarily focus on training the model with a more extensive collection of text-image paired data and employing stronger LLMs, significantly increasing training efforts. On the vision side, recent work concentrates on leveraging multiple vision encoders~\cite{lin2023sphinx, Gao2024SPHINXXSD} to enrich visual content, employing larger vision encoders~\cite{chen2023internvl}, and using advanced vision-language connectors~\cite{cha2023honeybee} to improve performance on multimodal tasks. However, these techniques result in an increased number of additional parameters and generate additional visual tokens for LLMs to process, making it inefficient to scale.

\begin{figure}[tb]
\centering
\includegraphics[width=0.48\textwidth]{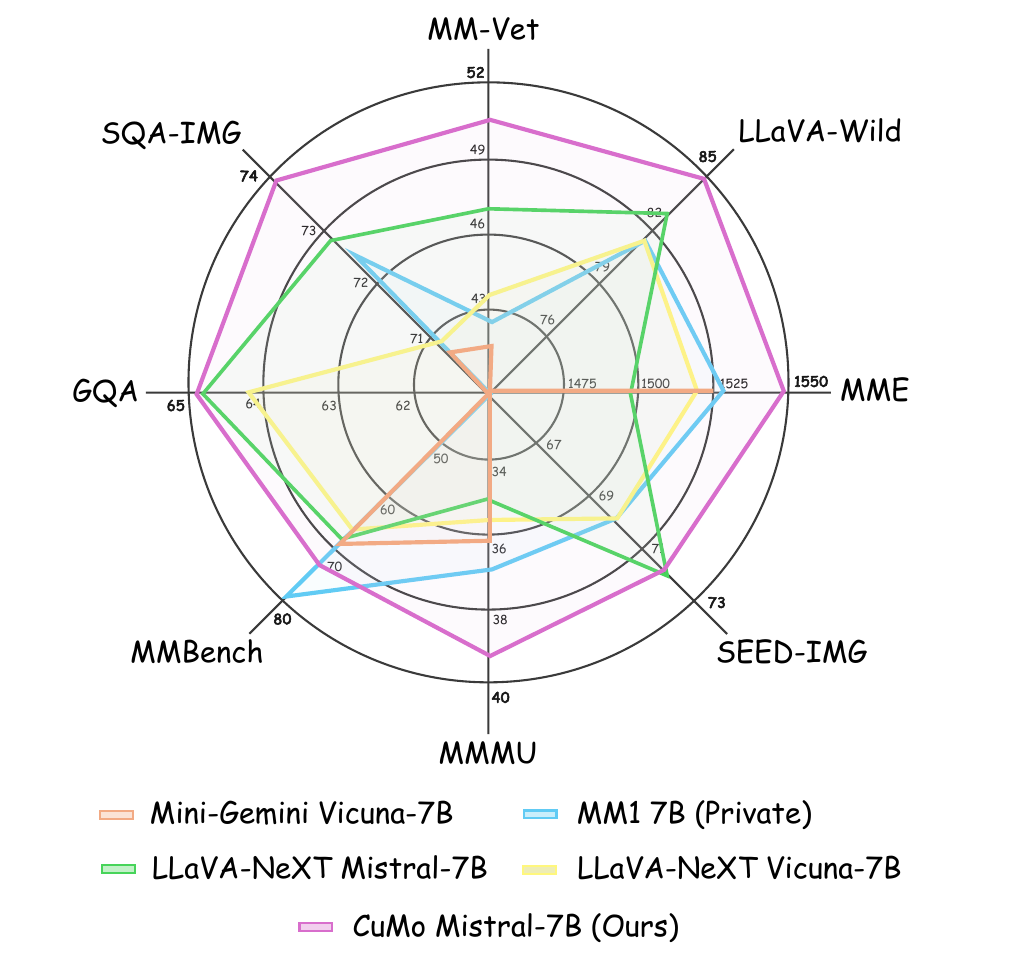}
\caption{Comparisons of CuMo Mistral-7B with state-of-the-art 7B multimodal LLMs. CuMo outperforms strong open-sourced models such as Mini-Gemini and LLaVA-NeXT, as well as the private MM1 model.}
\label{fig:teaser}
\end{figure}

\begin{figure*}[tb]
\centering
\includegraphics[width=0.98\textwidth]{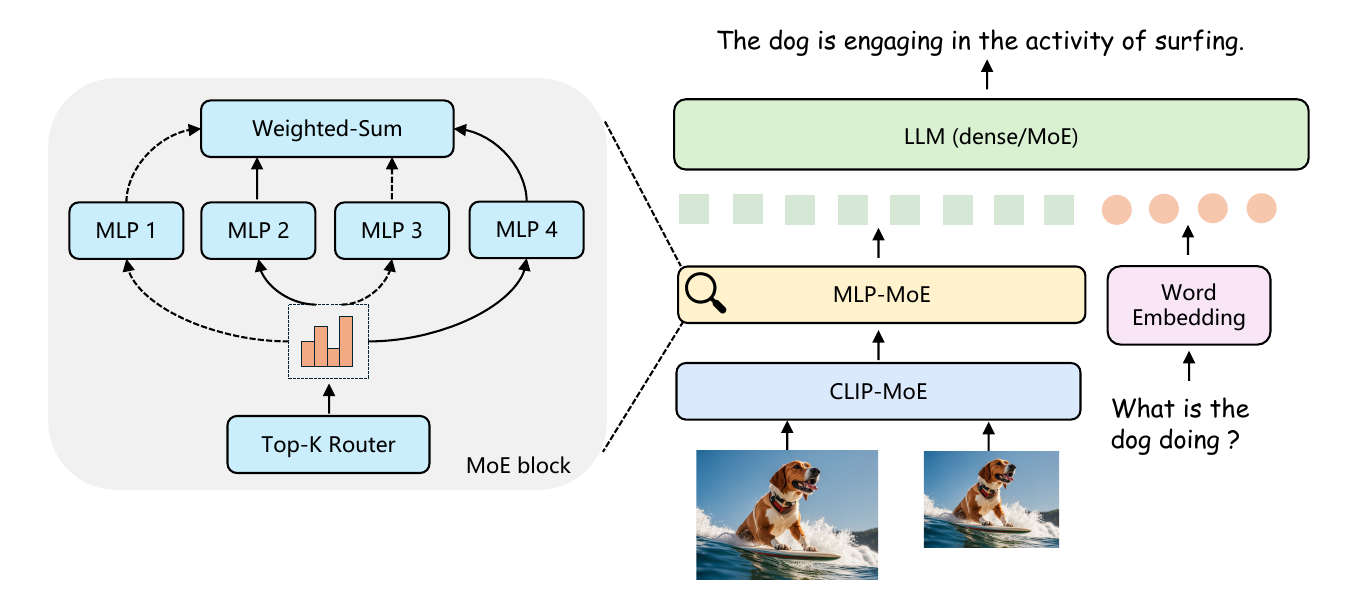}
\caption{\textbf{Architecture of CuMo.} CuMo incorporates sparse Top-K MoE blocks into the CLIP vision encoder and vision-language MLP connector, thereby improving the multimodal LLM capabilities
from the vision side. Skip connections are omitted for simplicity. Further implementation details are provided in Section~\ref{3.2}.}
\label{fig:archi}
\end{figure*}

In terms of efficiently scaling up models, Mixture-of-Experts (MoE) has become the de-facto framework in modern large-scale neural networks, particularly in natural language processing (NLP). Most large language models~(LLM) are built upon the transformer~\cite{vaswani2017attention} architecture, wherein sparse MoE is used to replace the dense MLP block with the Top-K sparsely-gated MoE block~\cite{shazeer2017outrageously}. Recent state-of-the-art open-sourced~\cite{mixtral, DBRX} and private~\cite{reid2024gemini} LLMs have predominantly adopted the sparse MoE architecture. These models are scaled up using the MoE design during training while maintaining relatively lower inference costs as only selected MLP experts are activated during the feed-forward process. Nevertheless, the development and optimization of MoE-based models have been largely tailored to LLMs, and the exploration of scaling multimodal LLMs with MoE, especially on the vision side, remains largely unexplored.

Motivated by these observations, we introduce CuMo, which integrates Top-K sparsely-gated MoE blocks into the vision encoder and the MLP connector of multimodal LLMs, as depicted in Figure~\ref{fig:archi}. We also explore the associated training recipe and methodology for CuMo. Firstly, we pre-train the MLP connector and perform pre-finetuning to warm up the whole model without introducing the MoE architecture, which stabilizes the following visual instruction tuning stage with newly incorporated sparse MoE blocks. Then, we replace each MLP block with the sparse MoE block in the MLP connector and the vision encoder through co-upcycling. Each expert within the sparse MoE block is initialized from the corresponding MLP block after the pre-training and the pre-finetuning stages. Additionally, each MoE block contains a Top-K router trained from scratch to select experts during the visual instruction tuning stage with auxiliary losses on the router to maintain a balanced loading of experts.
We conduct further comparisons between co-upcycled LLMs and pre-trained MoE-based LLMs. The results show that the pre-trained MoE-based LLMs significantly outperform the co-upcycled LLMs. As a result, the upcycling of LLMs is not included in CuMo.
Our models are trained fully on open-sourced datasets that are converted to visual instruction following formats. Experimental results demonstrate that CuMo outperforms other state-of-the-art multimodal LLMs on various VQA and multimodal instruction-following benchmarks within the same model size group, as illustrated in Figure~\ref{fig:teaser}.
Our contributions can be summarized as follows:
\begin{itemize}
    \item We introduce CuMo, which integrates co-upcycled sparsely-gated MoE layers into both the MLP connector and the vision encoder, enhancing the multimodal LLM with only slightly additional activated parameters.
    \item We outline the training methodology for CuMo, including a three-stage training process with auxiliary losses to stabilize training and ensure a balanced loading of experts.
    \item We train CuMo exclusively on open-sourced datasets and pre-trained models. It outperforms state-of-the-art open-sourced and private multimodal LLMs across multiple competitive benchmarks within each model size group.
    
\end{itemize}
\section{Related Works}
\subsection{Multimodal LLM}
While the ultimate goal for mulitmodal models may be generative across various modalities~\cite{xu2023versatile,bao2023one,tang2024any}, modern multimodal LLMs primarily focus on integrating additional modalities, such as vision, into LLMs.
InstructBLIP~\cite{dai2023instructblip} adopts Q-Former~\cite{li2023blip} to sample from visual tokens for LLM to feed-forward and follow the instructions. Flamingo~\cite{alayrac2022flamingo} and IDEFICS~\cite{idefics, laurenccon2024matters} use shared decoder for visual-language understanding. Qwen-VL~\cite{bai2023qwen} uses three-stage training to convert QwenLM to Qwen-VL. LLaVA series~\cite{liu2023llava, liu2023improvedllava, liu2024llavanext} adopt visual instruction tuning that uses instruction-following data to convert LLM into multimodal LLM. ShareGPT4V~\cite{chen2023sharegpt4v} collects detailed image caption data from GPT4V to augment the LLaVA models. HoneyBee~\cite{cha2023honeybee} investigates different designs of the MLP connector for better alignment. VILA~\cite{lin2023vila} unfreezes the LLM during pre-training with interleaved image-text data. MoE-LLaVA~\cite{lin2024moe} adopts the MoE design in small LLMs and reaches comparable performance to LLaVA with large LLMs. VCoder~\cite{jain2023vcoder} adopts various vision adapters to enhance visual perception abilities. SPHINX~\cite{lin2023sphinx, Gao2024SPHINXXSD} adopts multiple visual encoders to enrich the visual features with scaled data and models. InternLM-Xcomposer~\cite{zhang2023internlm, dong2024internlm} is trained with interleaved text-image composition data and achieves state-of-the-art performance. InternVL~\cite{chen2023internvl} scales up the vision encoder to a 6B ViT model. MM1~\cite{mckinzie2024mm1} summarizes the essential steps towards building a strong multimodal LLM from a pre-trained LLM. Mini-Gemini~\cite{li2024minigemini} further collects guided generation into the pipeline.

\subsection{Mixture-of-Experts}
Mixture-of-Experts~\cite{jacobs1991adaptive} is proposed to utilize a set of expert networks to address specific tasks by employing a gating network to determine the selection of these experts. Recently, it has gained popularity in the design of large language models~\cite{fedus2022review}. The mainstream practice~\cite{shazeer2017outrageously} is to replace the dense MLP layers with Top-K sparsely-gated mixture-of-experts~(MoE) layers in the transformer~\cite{vaswani2017attention}. \\
\noindent \textbf{MoE in Language} Subsequent works~\cite{Lepikhin2020GShardSG, fedus2022switch} have further scaled up MoE-based large language models with improved stability and load balancing of experts. The design of gating networks often involves selecting the top-k experts for each token~\cite{shazeer2017outrageously, Lepikhin2020GShardSG}. Various routing strategies have been explored, such as choosing top-k tokens by experts~\cite{zhou2022mixtureofexperts}, one-to-one matching between experts and tokens~\cite{lewis2021base}. Besides routing strategies, maintaining the load balance of experts is crucial for training MoE models. ST-MoE~\cite{zoph2022stmoe} adopts loading balancing loss and router-z loss to ensure a balanced distribution of the experts. Upcycling~\cite{komatsuzaki2023sparse} proposes training sparse experts from dense checkpoints to stabilize training and lower the cost. Recent large language models like Gemini-Pro~\cite{reid2024gemini} and DBRX~\cite{DBRX} are also based on the MoE design. \\
\noindent \textbf{MoE in Vision} The success of MoE extends to the vision community, particularly following the popularity of vision transformers~\cite{dosovitskiy2020image, carion2020end, zhu2020deformable,hassani2023neighborhood,hassani2022dilated,jain2023oneformer,li2024vmformer}. V-MoE~\cite{riquelme2021scaling} reaches comparable performance to dense ViT while only requiring half of the compute. LIMoE~\cite{mustafa2022multimodal} replaces dense MLP layers with MoE layers in CLIP and observes improvements in zero-shot image classification. Residual MoE~\cite{wu2022residual} corporates residual design into MoE transformer and saves over 30\% training cost. AdaMV-MoE~\cite{chen2023adamvmoe} proposes an adaptive MoE framework for multi-task learning. 
\section{Method}

\begin{figure}[tb]
\centering
\includegraphics[width=0.5\textwidth]{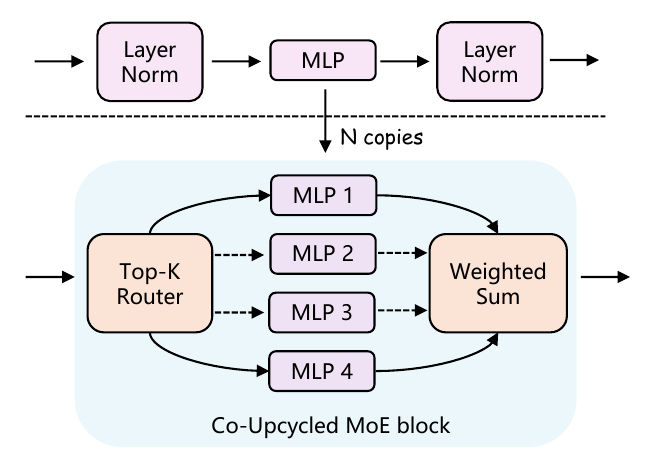}
\caption{\textbf{Initialization of MoE blocks via Co-Upcycling.} Each MLP expert within the MoE block during the visual instruction tuning stage is initialized from the corresponding pre-trained MLP.}
\label{fig:coupcycle}
\end{figure}

\begin{figure*}[tb]
\centering
\includegraphics[width=1.0\textwidth]{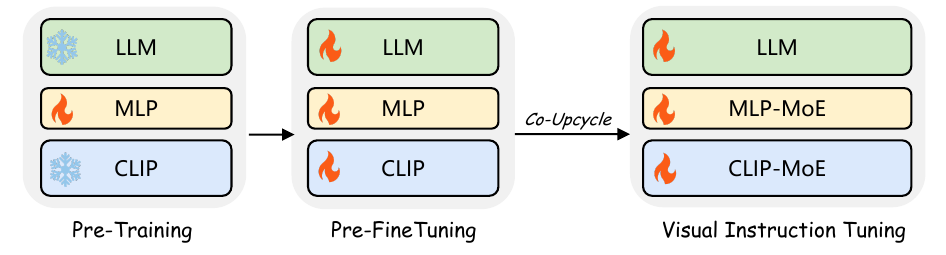}
\caption{\textbf{Training Stages of CuMo.} The first stage involves pre-training the MLP for better alignment. Subsequently, the pre-finetuning stage trains all parameters as a warm-up before the next stage. Finally, the MLP experts within each MoE block are initialized from the weights of the corresponding MLP block, followed by training all parameters in the visual instruction tuning stage.}
\label{fig:stage}
\end{figure*}

In this section, we first review the sparse MoE block structure and the upcycling strategy utilized in previous studies. Subsequently, we describe how these sparsely-gated MoE blocks are integrated into each module of multimodal LLMs using co-upcycling strategies. Then, we introduce the three-stage training process and auxiliary loss functions employed to stabilize training and balance the loads of experts.
\subsection{Revisit Sparse MoE}
\noindent \textbf{Sparse MoE Structure} Previous mainstream practice~\cite{shazeer2017outrageously} is to replace the dense MLP blocks with sparsely-gated mixture-of-experts blocks. Given input $\mathbf{X} \in \mathbb{R}^{N \times C_{in}}$ and a MLP block,
\begin{equation}
    X_{out} = \mathrm{MLP} (X) \in \mathbb{R}^{N \times C_{out}}
\end{equation} 
To scale up the model with multiple MLP blocks in parallel, a sparse MoE block includes a router network to select Top-K experts out of $S$ total experts. This router network has a linear layer to compute the normalized weight matrix based on the inputs $\mathbf{X}$ for voting, resulting in
\begin{equation}
    W = \mathrm{Softmax} (\mathrm{Linear} (X)) \in \mathbb{R}^{N \times S}
\end{equation} 
The Top-K experts are selected for each token based on $\mathbf{W}$, and the re-normalized weights $\mathbf{W_{K}} \in \mathbb{R}^{N \times K} $ are computed using
\begin{equation}
    W_{K} = \mathrm{Softmax} (\mathrm{TopK}(W)) \in \mathbb{R}^{N \times K}
\end{equation} 
Each selected expert is represented by an MLP block, and the final output is obtained through a re-weighted sum
\begin{equation}
    X_{out} = \sum_i^K W^i_{K} \circ \mathrm{MLP}_{i}(X) \in \mathbb{R}^{N \times C_{out}}
\end{equation}
the output $\mathbf{X}_{out}$ maintains the same dimension as the output of a single dense MLP block.

\noindent \textbf{Sparse Upcycling} Training MoE-based designs from scratch can be unstable and costly. Sparse Upcycling~\cite{komatsuzaki2023sparse} addresses this challenge by initializing the experts in each MoE block from the corresponding MLP block in pre-trained dense checkpoints. This initialization approach provides a better starting point for training MoE-based models and reduces training costs compared to training from scratch.

\subsection{CuMo Architecture}
\label{3.2}
\noindent \textbf{Sparse MoE in MLP Connector} The MLP connector converts visual tokens into word embedding space, aligning dimensions between visual and text tokens. An effective architecture for the vision-language connector is an MLP block~\cite{liu2023improvedllava} that contains two linear layers. We start from a single MLP block and replace it with a Top-K sparse MoE block, incorporating a Top-K router and a set of experts for projecting visual tokens into word embedding space.

\noindent \textbf{Sparse MoE in Vision Encoder} Vision encoders extract image features as sequences of visual tokens for reasoning in LLMs. CLIP~\cite{CLIP} is one the most popular pre-trained vision encoders for multimodal LLM since it is pre-trained on large-scale image-text pairs, which makes it suitable for processing images for multimodal usage. The visual encoding part of CLIP is a ViT~\cite{dosovitskiy2020image} model, which has consecutive MLP blocks in the transformer encoder. We substitute each MLP block with a Top-K sparse MoE block, retaining skip connections alongside MoE block outputs.

\noindent \textbf{Sparse MoE in LLM} In terms of using MoE in LLM, we compare the co-upcycled LLM with pre-trained MoE-based LLM. We start from Mistral-7B and the upcycled Mistral-7B-MoE slightly outperforms Mistral-7B on certain benchmarks. However, considering the constrained knowledge base of upcycled experts from Mistral-7B, we compare it with the pre-trained Mixtral 8x7B with pre-trained experts of a diverse knowledge base. Experimental results reveal that pre-trained Mixtral 8x7B significantly outperforms Mistral-7B-MoE. As a result, LLM is not co-upcycled with CLIP and MLP connectors since it brings marginal improvements with great additional parameters.

\begin{table*}[tb]
\centering
\resizebox{1.0\textwidth}{!}{
\begin{tabular}{l|l|c|ccccccccccccc}
&  & &SQA &Text & & & &\multicolumn{2}{c}{MMB} &MM &VQA &LLaVA &SEED &MMMU &Math \\  
Method  &LLM  &Act. &IMG &VQA &GQA &POPE &MME &EN &CN &Vet &v2 &Wild &IMG &val &Vista \\ \midrule
\multicolumn{2}{c}{\textit{7B to 13B Models}} \\ \midrule
InstructBLIP~\cite{dai2023instructblip} &Vicuna-7B &7.9B &60.5 &50.1 &49.2 &- &- &36.0 &23.7 &26.2 &- &60.9 &60.5 &- &- \\
Qwen-VL-Chat~\cite{bai2023qwen} &Qwen-7B &- &68.2 &61.5 &57.5 &- &1487.5 &60.6 &56.7 &-  &78.2 &- &58.2 &35.9 &- \\
LLaVA-v1.5~\cite{liu2023improvedllava} &Vicuna-7B &7.1B &66.8 &58.2 &62.0 &85.9 &1510.7 &64.3 &58.3 &30.5 &78.5 &63.4 &66.1 &- &- \\
LLaMA-VID~\cite{llamavid} &Vicuna-7B &- &68.3 &- &64.3 &86.0 &1521.4 &65.1 &- &- &79.3 &- &59.9 &- &- \\
VILA~\cite{lin2023vila} &Vicuna-7B &7.1B &68.2 &64.4 &62.3 &85.5 &1533.0 &68.9 &61.7 &34.9 &79.9 &69.7 &61.1 &- &- \\
SPHINX-Intern2~\cite{Gao2024SPHINXXSD} &InternLM2-7B  &- &70.4 &58.1 &56.2 &86.9 &1260.4 &57.9 &- &36.5 &75.5 &57.6 &68.8 &- &35.5 \\
LLaVA-NeXT~\cite{liu2024llavanext}  &Mistral-7B &7.6B &72.8 &65.7 &64.8 &86.7 &1498	 &68.7 &61.2 &47.3 &82.2 &83.2	&72.2 &35.3	&37.7 \\
LLaVA-NeXT~\cite{liu2024llavanext}  &Vicuna-7B &7.1B &70.1 &64.9 &64.2 &86.5 &1519 &67.4 &60.6 &43.9 &81.8 &81.6 &70.2 &35.8 &34.6 \\
LLaVA-LLaMA3~\cite{2023xtuner}  &LLaMA3-8B-IT &8.4B &72.9 &59.0 &62.6 &86.4 &1469 &72.3 &66.4 &- &- &- &70.1 &36.8 &- \\ 
Mini-Gemini~\cite{li2024minigemini}  &Vicuna-7B  &7.3B &65.2 &- &- &- &1523 &69.3 &- &40.8 &- &- &- &36.1 &31.4 \\
MM1~\cite{mckinzie2024mm1} &MM1-7B &- &72.6 &72.8 &- &86.6 &1529.3 &79.0 &- &42.1  &82.8 &81.5 &69.9 &37.0 &35.9 \\ \midrule
InstructBLIP~\cite{dai2023instructblip} &Vicuna-13B &14.2B &63.1 &50.7 &49.5 &78.9 &1212.8 &- &-  &25.6 &- &58.2 &63.1 &- &- \\
LLaVA-v1.5~\cite{liu2023improvedllava} &Vicuna-13B &13.4B &71.6 &61.3 &63.3 &85.9 &1531.3 &67.7 &63.6 &35.4 &80.0 &70.7 &68.2 &36.4 &27.6 \\
VILA~\cite{lin2023vila} &Vicuna-13B &13.4B &73.7 &66.6 &63.3 &84.2 &1570.1 &70.3 &64.3 &38.8 &80.8 &73.0 &62.8 &- &- \\
LLaMA-VID~\cite{llamavid} &Vicuna-13B &- &70.0 &- &65.0 &86.0 &1542.3 &66.6 &- &- &80.0 &- &62.3 &- &- \\
SPHINX-Plus~\cite{Gao2024SPHINXXSD}  &LLaMA2-13B &- &74.2 &65.7 &- &89.1 &1457.7 &71.0 &- &47.9 &- &71.7 &74.8 &- &36.8 \\
Mini-Gemini\cite{li2024minigemini}  &Vicuna-13B &13.6B &65.9 &- &- &- &1565 &68.5 &- &46.0 &- &- &- &38.1 &37.0 \\
InternVL-Chat~\cite{chen2023internvl} &Vicuna-13B &19B &- &61.5 &66.6 &87.6 &1586.4 &- &- &- &81.2 &- &- &- &- \\
LLaVA-NeXT~\cite{liu2024llavanext} &Vicuna-13B &13.4B &73.6 &67.1 &65.4 &86.2 &1575 &70 &64.4 &48.4 &82.8 &87.3 &71.9 &36.2 &35.3 \\  \midrule
\rowcolor{lightblue}
CuMo &Mistral-7B &7.8B &73.9 &67.0 &64.9 &86.7 &1548.6 &73.0 &66.6 &51.0\textsuperscript{\textdagger} &82.2 &85.7\textsuperscript{\textdagger} &72.1 &39.1  &35.1\textsuperscript{\textdagger} \\\midrule
\multicolumn{2}{c}{\textit{7B MoE Models}} \\ \midrule
SPHINX-MoE~\cite{Gao2024SPHINXXSD} &Mixtral-8$\times$7B &- &74.5 &68.0 &63.8 &89.6 &1485.3 &71.3 &- &40.9 &81.1 &70.2 &73.0 &31.1 &42.7  \\ 
MM1~\cite{mckinzie2024mm1} &MM1-7B-MoE &- &75.3 &72.8 &- &87.6 &1629.0 &79.7 &- &47.0 &83.4 &82.0 &70.4 &40.9 &40.9 \\ 
Mini-Gemini~\cite{li2024minigemini}  &Mixtral-8$\times$7B &13.5B &- &69.2 &- &- &1639 &75.6 &- &45.8 &- &- &- &41.8 &41.8 \\ \midrule
\rowcolor{lightblue}
CuMo &Mixtral-8$\times$7B  &13.5B &77.9 &66.0 &63.8 &85.7 &1639.5 &75.3 &68.0 &48.7\textsuperscript{\textdagger} &81.8  &84.7\textsuperscript{\textdagger} &73.2 &45.0 &38.2\textsuperscript{\textdagger} \\ \midrule
\multicolumn{2}{c}{\textit{Private Models}} \\ \midrule
GPT4V~\cite{gpt4v} &- &- &- &78.0 &-  &- &- &77.0 &74.4 &60.2 &- &- &- &56.8 &49.9 \\
Gemini 1.5 Pro~\cite{reid2024gemini} &- &- &- &73.5  &-  &- &- &73.6 &74.3 &64.3 &73.2 &- &- &58.5 &52.1 \\
Claude 3 Opus~\cite{claude3} &- &- &- &-  &-  &- &- &63.3 &59.2 &58.1 &- &- &- &59.4 &50.5 \\
Qwen-VL-Max~\cite{qwenvlmax} &- &- &- &79.5  &-  &- &1790.1 &77.6 &75.1 &66.6 &- &- &- &51.4 &51.0 \\
\midrule
\end{tabular}}
\caption{Comparisons between CuMo and other state-of-the-art multimodal LLMs on competitive benchmarks. These models are grouped by the size of the base LLM. The benchmarks are double-rowed due to limited space: SQA-IMG~\cite{scienceqa}; TextVQA~\cite{textvqa}; GQA~\cite{gqa}; POPE~\cite{pope}; MME~\cite{mme}; MMBench~\cite{mmbench}; MMVet~\cite{mmvet}; VQAv2~\cite{vqav2}; LLaVA-Wild~\cite{liu2023llava}; SEED-IMG~\cite{seed}; MMMU~\cite{mmmu}; MathVista~\cite{mathvista}. Act.: Activated Parameters. Numbers\textsuperscript{\textdagger} are averaged by three inference runs of querying GPT API.} 
\label{tab:m3_final}
\end{table*}

\subsection{Training Recipe}
\label{3.3}
\noindent \textbf{Co-Upcycling MoE blocks} We start with training the added MoE blocks from scratch while the model is struggling to converge. Attempts to address this issue with lower learning rates perform worse compared to the baseline. As a result, we adopt a co-upcycling approach, initializing each module that integrates sparsely-gated MoE blocks with pre-trained MLPs to replace corresponding MLP blocks, as shown in Figure~\ref{fig:coupcycle}. This strategy consistently improves training stability and model performance.

\noindent \textbf{Three-Stage Training} To further enhance training stability, we adopt a three-stage training strategy for CuMo models, as illustrated in Figure~\ref{fig:stage}. In the first stage, we only pre-train the MLP connector, given that the vision encoder and LLM have already undergone pre-training on large-scale data. During the second pre-finetuning stage, we train all parameters using high-quality caption data to warm up the entire model before introducing MoE blocks in the subsequent stage. The third stage involves visual instruction finetuning, where the multimodal LLM is scaled up with upcycled MoE blocks and trained on visual instruction tuning data.

\noindent \textbf{Loss Function} To maintain a load balance between experts in each MoE block, we adopt auxiliary losses based on the language modeling cross-entropy loss. The auxiliary losses comprise loading balance loss and router z-loss~\cite{zoph2022stmoe}. Hence, the total loss is 
\begin{equation}
    L = L_{ce} + \alpha_b L_{b} + \alpha_z L_{z}
\end{equation}
Here, $L_{ce}$ represents the language modeling loss, which computes the cross-entropy of next-token predictions. $\alpha_b$ and $\alpha_z$ denote coefficients for loading balance loss $L_{b}$ and router z-loss $L_{z}$, set to 0.1 and 0.01, respectively, across all experiments. These auxiliary losses, abbreviated as bzloss in Section~\ref{experiments}, are individually applied to the MLP connector, vision encoder, and LLM for simplicity.

\section{Experiments}
\label{experiments}

We train the CuMo models on a mixture of open-sourced datasets, which are converted into the visual instruction tuning format. Then, we conduct comprehensive evaluations of the performance of CuMo models across various competitive VQA-based and instruction-following-based benchmarks. Additionally, we perform ablation studies on each module with upcycled MoE blocks with qualitative analysis of the results.

\begin{table*}[tb]
\centering
\resizebox{1.0\textwidth}{!}{
\begin{tabular}{l|l|cc|ccccccccccc}
& & & &SQA &Text & & & &\multicolumn{2}{c}{MMBench} &MM &VQA &LLaVA &SEED \\ 
Method &LLM &PT &IT &IMG &VQA &GQA &POPE &MME &EN &CN &Vet  &v2 &Wild &IMG \\ \midrule
InstructBLIP~\cite{dai2023instructblip} &Vicuna-7B &129M &1.2M &60.5 &50.1 &49.2 &- &- &36.0 &23.7 &26.2 &- &60.9 &60.5 \\
InstructBLIP~\cite{dai2023instructblip} &Vicuna-13B &129M &1.2M &63.1 &50.7 &49.5 &78.9 &1212.8 &- &-  &25.6 &- &58.2 &63.1 \\
IDEFICS-9B~\cite{idefics} &LLaMA-7B &353M &1M &- &25.9 &38.4 &- &- &48.2 &25.2  &-  &50.9 &- &-  \\ 
IDEFICS-80B~\cite{idefics} &LLaMA-65B &353M &1M &- &30.9 &45.2 &- &- &54.5 &38.1 &-  &60.0 &- &- \\
Qwen-VL~\cite{bai2023qwen} &Qwen-7B &1.4B &50M &67.1 &63.8 &59.3 &- &- &38.2 &7.4 &-  &78.8 &- &56.3 \\
Qwen-VL-Chat~\cite{bai2023qwen} &Qwen-7B &1.4B &50M &68.2 &61.5 &57.5 &- &1487.5 &60.6 &56.7 &-  &78.2 &- &58.2\\
LLaVA-v1.5~\cite{liu2023improvedllava}  &Vicuna-7B &558K &665K &66.8 &58.2 &62.0 &85.9 &1510.7 &64.3 &58.3 &30.5 &78.5 &63.4 &66.1 \\
LLaVA-v1.5~\cite{liu2023improvedllava} &Vicuna-13B &558K &665K &71.6 &61.3 &63.3 &85.9 &1531.3 &67.7 &63.6 &35.4 &80.0 &70.7 &68.2 \\ \midrule
\rowcolor{lightblue}
CuMo &Mistral-7B &558K &665K &71.7 &59.3 &63.2 &87.1 &1428.6 &69.6 &62.6 &34.3 &80.6 &68.8 &69.6 \\ \midrule
\end{tabular}}
\caption{Comparisons between CuMo Mistral-7B and other multimodal LMM models with limited training data.}
\label{tab:m3_665k}
\end{table*}
\begin{table}[tb]
\resizebox{0.5\textwidth}{!}{
\begin{tabular}{l|cccc}
Method &SQA &VQA$^T$ &MMVet &SEED \\ \hline
Baseline on Mistral-7B   &72.8 &57.6 &32.1 &66.4   \\ \hline
+ \textit{Top 2-in-4 \& Scratch} &68.1 &55.6 &29.3 &65.1 \\
\rowcolor{lightblue}
$\rightleftharpoons$ \textit{Top 2-in-4 \& Upcycle} &73.7 &57.2 &32.3 &67.1 \\
\rowcolor{lightblue}
+ \textit{bzloss}  &73.5 &57.4 &33.1 &67.4 \\
$\rightleftharpoons$ \textit{Top 2-in-8 \& Upcycle}  &73.4 &57.6 &32.4 &67.2 \\\hline
\end{tabular}}
\caption{Ablation study on the MLP-MoE module. Each row represents a different configuration, with changes or additions marked using $\rightleftharpoons$ and $+$ symbols, respectively. Settings highlighted with a light blue background are those adapted for the MLP-MoE module in Table \ref{tab:m3_final}.}
\label{tab:m3_ablation_mlpsmoe}
\end{table}
\begin{table}[tb]
\centering
\resizebox{0.5\textwidth}{!}{
\begin{tabular}{l|ccccc}
Method &SQA &VQA$^{T}$ &MMVet &SEED \\ \hline
MLP-MoE  &73.5 &57.4&33.1 &67.4 \\
+ \textit{Unfreeze CLIP} &72.0 &58.9 &34.7 &69.0 \\ \hline
\rowcolor{lightblue}
+ \textit{Top 2-in-4 \& bzloss} &72.8 &59.7 &35.4 &69.8  \\
$\rightleftharpoons$ \textit{Top 2-in-8 \& bzloss} &71.0 &59.0 &33.6 &69.2 \\

\end{tabular}}
\caption{Ablation study on the CLIP-MoE module. All MoE blocks in CLIP are initialized with upcycling.}
\label{tab:m3_ablation_clipsmoe}
\end{table}
\begin{table}[tb]
\centering
\resizebox{0.5\textwidth}{!}{
\begin{tabular}{l|cccc}
Method & SQA &VQA$^T$ &MMVet &SEED \\ \hline
MLP-MoE \& CLIP-MoE &71.7 &59.3 &34.3 &69.6 \\ \hline
+ \textit{Mistral 4$\times$7B \& Upcycle} &72.8 &57.0 &35.2 &69.9 \\
$\rightleftharpoons$ \textit{Mistral 8$\times$7B \& Upcycle} &73.2 &56.4 &35.7 &70.5 \\
\rowcolor{lightblue}
$\rightleftharpoons$ \textit{Mixtral 8$\times$7B} &74.2 &60.6 &40.0 &72.6 \\
\end{tabular}}
\caption{Ablation study on the LLM-MoE module. Mixtral 8$\times$7B outperforms upcycled Mistral MoE models significantly.}
\label{tab:m3_ablation_llmsmoe}
\end{table}

\subsection{Implementation Details}
\label{4.1}
\noindent \textbf{Training Datasets} During pre-training, we only utilize LLaVA-558K~\cite{liu2023llava} to train the MLP connector for better alignment. In the subsequent pre-finetuning stage, detailed image caption data from ALLaVA~\cite{chen2024allava} is employed to warm up all parameters of the multimodal LLM. For the final visual instruction tuning stage, a mixture of datasets including LLaVA-665K~\cite{liu2023improvedllava}, ShareGPT4V~\cite{chen2023sharegpt4v}, LAION-GPT-V~\cite{gpt4vdataset}, DocVQA~\cite{docvqa}, ChartQA~\cite{chartqa}, AI2D~\cite{ai2d}, InfoVQA~\cite{mathew2022infographicvqa}, SynDog-EN~\cite{kim2022ocr}, ALLaVA~\cite{chen2024allava}, and LIMA~\cite{zhou2024lima} is utilized to train the CuMo models with upcycled MoE blocks. The total data size for visual instruction tuning is approximately 1.65 million, and all training data are publicly accessible. The detailed breakdown of the training dataset is listed in Appendix~\ref{appendix:dataset}.

\noindent \textbf{Evaluation Benchmarks} Evaluation of CuMo models primarily focuses on academic VQA-based datasets such as VQAv2~\cite{vqav2}, GQA~\cite{gqa}, Science-QA~\cite{scienceqa}, and TextVQA~\cite{textvqa}, as well as instruction-following-based LMM benchmarks including POPE~\cite{pope}, MME~\cite{mme}, MMBench~\cite{mmbench}, SEED-Bench~\cite{seed}, LLaVA-Wild~\cite{liu2023llava}, and MM-Vet~\cite{mmvet}. Additionally, the challenging MMMU~\cite{mmmu} and MathVista~\cite{mathvista} datasets are evaluated to assess the visual reasoning abilities of the multimodal LLMs.

\noindent \textbf{Training Settings} We employ the pre-trained CLIP ViT-L~\cite{CLIP} as the vision encoder, a two-layer MLP as the vision-language connector, and Mistral-7B~\cite{mistral} as the LLM to establish the baseline model following LLaVA v1.5~\cite{liu2023improvedllava}. We only use LLaVA-558K~\cite{liu2023improvedllava} as pre-training data and LLaVA-665K~\cite{liu2023improvedllava} as visual instruction tuning data to train the baseline model and make ablation studies for comparisons. The learning rate is set to 1e-3 for pre-training the MLP connector and reduced to 2e-5 for visual instruction tuning of both the MLP connector and CLIP. To further stabilize the visual instruction tuning process after scaling up with additional data, the learning rate is lowered to 2e-6 for all parameters of the CuMo models in the final results. More hyperparameters of the training process is listed in Appendix~\ref{appendix:setup}.

\noindent \textbf{Evaluation Settings} During evaluation, we adhere to the settings outlined in the LLaVA series~\cite{liu2023improvedllava}, employing a greedy decoding strategy for all benchmarks. The data and questions are converted into visual instructions to prompt the multimodal LLMs. For benchmarks that utilize GPT API for evaluation, we adopt gpt-4-0613 for LLaVA-Wild~\cite{liu2023llava} and gpt-3.5-turbo for MathVista~\cite{mathvista}.

\subsection{Main Results}
\noindent \textbf{Comparison with SoTA Multimodal LLMs} In Table~\ref{tab:m3_final}, we present a comparison of CuMo models with other state-of-the-art instruction-following-based multimodal LLMs. We categorize the models based on the size of the base LLMs, including 7B models, 13B models, and 7B MoE models. CuMo Mistral-7B outperforms other 7B-based state-of-the-art multimodal LLMs across multiple benchmarks. Moreover, the performance of the CuMo Mistral-7B model is comparable to many 13B-based multimodal LLMs. In the case of Mixtral-8$\times$7B models, CuMo achieves results on par with SPHINX-MoE, MM1, and Mini-Gemini. LLaMA-based LLMs~\cite{chiangvicuna, touvron2023llama} are not utilized in our experiments due to license constraints.

\noindent \textbf{Comparison under limited training data} To further evaluate the effectiveness of the co-upcycled MoE blocks, we train the vanilla CuMo mistral-7B under limited training data in Table~\ref{tab:m3_665k}. It shows that CuMo outperforms other 7B models and reaches comparable performance to LLaVA-v1.5 Vicuna-13B under the same training data.

\begin{table}[tb]
\centering
\resizebox{0.5\textwidth}{!}{
\begin{tabular}{ccc|cccc}
$1\times$ &$2\times$ &$3\times$ &SQA &VQA$^{T}$ &MMVet  &SEED  \\ \hline
\checkmark &- &- &71.7 &59.3 &34.3 &69.6 \\ \hline
\checkmark &\checkmark &- &71.7 &60.6  &35.0 &69.7 \\
\rowcolor{lightblue}
\checkmark &- &\checkmark  &72.9 &61.0 &37.0 &69.7\\
\checkmark &\checkmark &\checkmark  &72.2 &60.5 &36.9 &70.1 \\\hline
\end{tabular}}
\caption{Ablation study on multi-resolution image features. The combination of $3\times$ and $1\times$ is adopted for the final models in Table~\ref{tab:m3_final}.}
\label{tab:m3_ablation_ms}
\end{table}
\begin{table}[tb]
\centering
\resizebox{0.5\textwidth}{!}{
\begin{tabular}{l|cccc}
Method & SQA &VQA$^{T}$ &MMVet  &SEED \\ \hline
No PFT &71.7 &59.3 &34.3  &69.6 \\ \hline
+ \textit{ShareGPT4V} &72.4 &61.7 &36.5 &70.0 \\
\rowcolor{lightblue}
$\rightleftharpoons$ \textit{ALLaVA} &73.0 &62.8 &37.2 &70.9 \\
\end{tabular}}
\caption{Ablation study on the pre-finetuning stage. ALLaVA is chosen for pre-finetuning due to its provision of high-quality image caption data.}
\label{tab:m3_ablation_pft}
\end{table}

\subsection{Ablation Study}
\label{4.3}
\noindent \textbf{Upcycle MLP connector to MLP-MoE} We initiate the ablation study by replacing the MLP connector with upcycled MLP-MoE, as depicted in Table~\ref{tab:m3_ablation_mlpsmoe}. We start with a Top 2-in-4 router and train the MoE blocks from scratch, which leads to a clear performance drop on all benchmarks. Then, we adopt the upcycling strategy to initialize the MLP experts. We observe marginal improvements over the baseline, considering each expert comprises only two linear layers. Subsequently, the incorporation of bzloss to ensure a balanced loading of experts in the MLP-MoE yields noticeable enhancements on MMVet. However, employing a Top 2-in-8 router with upcycling and bzloss results in a slight performance decline, possibly due to the limited visual instruction tuning data to train robust and well-balanced eight experts.

\noindent \textbf{Empower CLIP with CLIP-MoE} In Table~\ref{tab:m3_ablation_clipsmoe}, initially unfreezing CLIP based on MLP-MoE leads to noticeable improvements on TextVQA and MMVet benchmarks. However, training the added Top2-in-4 MoE blocks in CLIP from scratch proves unsuccessful, as the model fails to converge even with reduced learning rates. Consequently, adopting upcycled MoE blocks during the visual instruction tuning stage yields further enhancements on TextVQA, MMVet, and SEED benchmarks.

\begin{figure}[tb]
\centering
\includegraphics[width=0.5\textwidth]{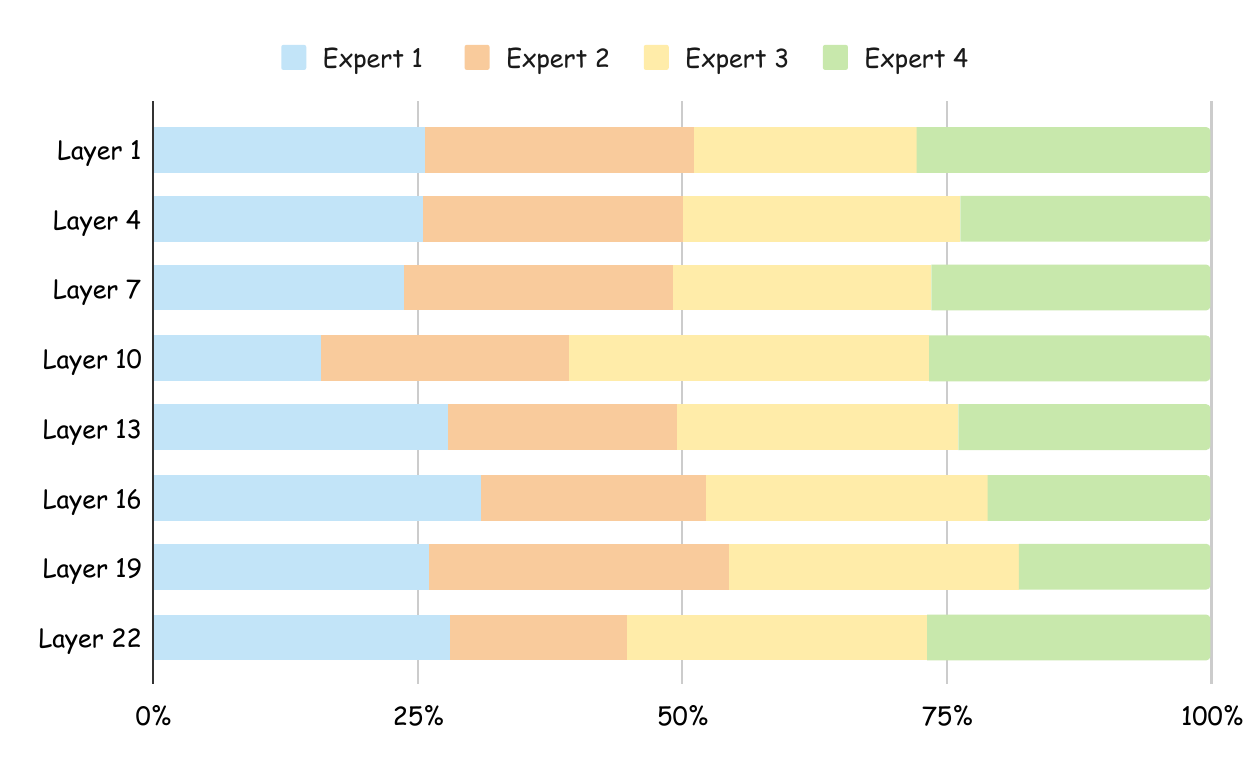}
\caption{\textbf{Expert distributions of MoE blocks in CLIP.} We select layers from CLIP and summarize the activated experts during the feed-forward process on the MME test set.}
\label{fig:expert_distribution}
\end{figure}

\begin{figure*}[tb]
\centering
\includegraphics[width=1.0\textwidth]{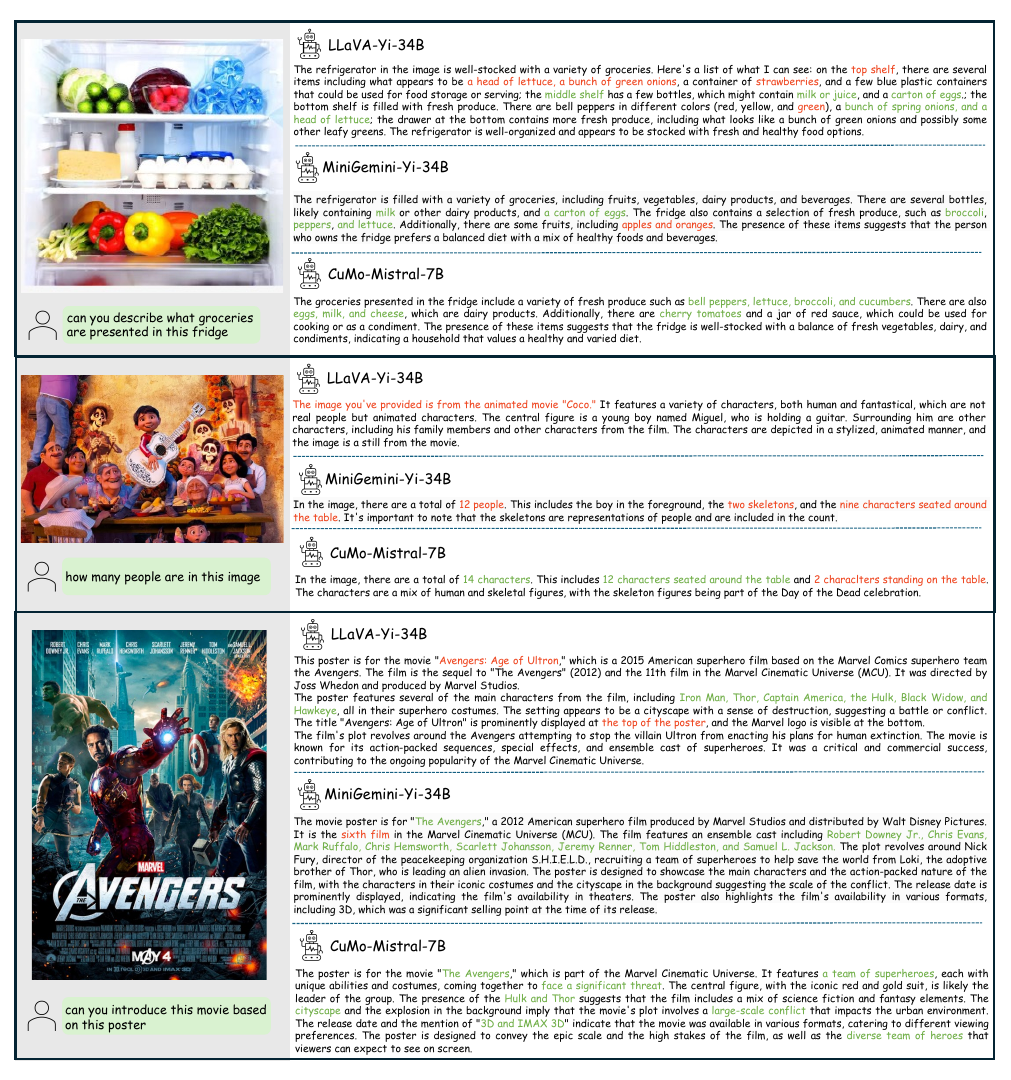}
\caption{\textbf{Dialogues between the user and multimodal LLMs on challenging images.} We highlight the \textcolor{demogreen}{correct} answers and \textcolor{demored}{hallucinations} from the responses of the multimodal LLMs.}
\label{fig:demo}
\end{figure*}

\noindent \textbf{Upcycle LLM vs Pre-trained LLM-MoE} 
Upon replacing all MLP blocks with sparsely-gated MoE blocks in the visual part, we further investigate the utilization of the MoE architecture in the LLM.
Starting from the Mistral-7B model, we first lower the learning rate to 2e-6 to set the baseline and the following experiments since a learning rate of 2e-5 induces training instabilities.
Then, we upcycle each MLP block with a sparsely-gated MoE block, initializing the weight of each expert from the pre-trained MLP block. As demonstrated in Table~\ref{tab:m3_ablation_llmsmoe}, the upcycled Mistral-4$\times$7B and 8$\times$7B outperform the Mistral-7B model slightly except for TextVQA. However, considering that the upcycled experts significantly increase parameters without introducing new knowledge, we replace the upcycled Mistral 8$\times$7B with Mixtral 8$\times$7B~\cite{mixtral}. In Mixtral 8$\times$7B, all expert layers are pre-trained on large-scale language data, providing superior initialization compared to upcycling. The results indicate that CuMo Mixtral-8x7B outperforms its upcycled counterparts significantly and is employed in the final models with bzloss to maintain a balanced loading of experts.

\noindent \textbf{Multi-Resolution Visual Features} Incorporating multi-resolution inputs is crucial for enhancing the understanding of image content in multimodal LLMs. Following the approach outlined in $S^2$\cite{shi2024we}, we introduce multi-resolution inputs to CLIP and concatenate the feature maps channel-wise to maintain the total number of visual tokens consistent with low-resolution inputs. As illustrated in Table~\ref{tab:m3_ablation_ms}, an empirical combination of 3$\times$ and 1$\times$ yields optimal performance and we adopt this configuration for the final CuMo models.

\noindent \textbf{Pre-FineTuning Stage} Previous ablation studies were conducted directly after the pre-training of the MLP connector, leading to observed training instabilities during visual instruction tuning. To address this, we introduce a pre-finetuning stage using high-quality image caption data, wherein all parameters are unfrozen. In Table~\ref{tab:m3_ablation_pft}, we leverage caption data from ALLaVA for this stage. Results indicate that ALLaVA data proves to be a superior option, providing fewer but higher-quality captions for training, ultimately leading to improved performance.

\subsection{Qualitative Analysis}
\noindent \textbf{Expert Distribution} As shown in Figure~\ref{fig:expert_distribution}, we visualize the expert distributions in the MoE block from selected layers at CLIP-MoE. The dataset analyzed is the test set of the MME benchmark. The distribution indicates that the selected experts during inference are evenly spread across layers, providing further evidence of the effectiveness of the auxiliary losses in maintaining load balance.

\noindent \textbf{Dialogue Comparisons} Presented in Figure~\ref{fig:demo}, we contrast the responses from CuMo-Mistral-7B, LLaVA-Yi-34B, and MiniGemini-Yi-34B. It demonstrates that CuMo-Mistral-7B can effectively follow instructions and predominantly provide correct answers to challenging questions derived from complex scenes. However, CuMo also exhibits instances of hallucinations, such as responding with ``2 characters standing on the table," highlighting the need for further investigation to mitigate hallucinations in CuMo.

\section{Conclusion}
In this study, we introduce the sparse mixture-of-experts design into multimodal LLMs. Specifically, we replace each MLP block with a Top-K sparse MoE block in the MLP connector and the vision encoder. To enhance training stability, we employ a three-stage training approach, incorporating upcycled MoE blocks during the visual instruction tuning stage, along with auxiliary bzloss to maintain a balanced loading of experts. All CuMo models are trained and evaluated on fully open-sourced datasets and benchmarks. Through extensive experiments and ablation studies, we validate the effectiveness of the upcycled MoE blocks in each module. CuMo outperforms state-of-the-art models across multiple competitive benchmarks within the same group of model sizes.

\paragraph{Acknowledgments}
We extend our gratitude to Chunyuan Li, Lei Chen, and Haibin Lin for their insightful and valuable discussions throughout this project.


{
    \small
    \bibliographystyle{ieeenat_fullname}
    \bibliography{main}
}

\clearpage
\newpage
\appendix
\section*{\Large{Appendix}}
The supplementary material elaborates on further aspects of our work concerning the experimental setups and dataset usage.
In Appendix~\ref{appendix:dataset}, we provide details on the datasets used for the visual instruction tuning stage and how we converted the mixture of datasets into the visual instruction following formats. In Appendix~\ref{appendix:setup}, we present the hyperparameters used for the three-stage trainings. In Appendix~\ref{appendix:dialogues}, we include additional examples of dialogues between the user and our M3 models.

\begin{table}[tb]
\centering
\begin{tabular}{c|c}
Dataset &Size \\ \hline
\multicolumn{2}{c}{\textit{Pre-Training}} \\ \hline
LCS-558K &558K  \\ \hline
\multicolumn{2}{c}{\textit{Pre-Finetuning}} \\ \hline
ALLaVA-Caption &708K   \\ \hline
\multicolumn{2}{c}{\textit{Visual Instruction Tuning}} \\ \hline
LLaVA-665K &665K \\
ShareGPT4V &102K \\
LAION-GPT-V &11K  \\
DocVQA &10K \\
SynDog-EN &50K \\
ChartQA  &4K  \\
DVQA & 50K \\
AI2D &2K \\
InfoVQA & 4K \\
ALLaVA &708K  \\
LIMA &1K  \\
ALLaVA-Text &143K  \\
\hline
\end{tabular}
\caption{List of datasets used for three training stages.} 
\label{tab:app_datasets}
\end{table}

\section{Dataset Details}
\label{appendix:dataset}
As outlined in Table~\ref{tab:app_datasets}, we provide detailed information on the datasets utilized for the three-stage training process mentioned in Section~\ref{3.3}. All data are converted into the instruction-following format for training. For the Syndog-EN and DVQA datasets, we didn't use the entire training set as we observed that a large portion of synthetic data negatively impacts the zero-shot performance of the multimodal LLMs.

\begin{table}[tb]
\centering
\resizebox{0.5\textwidth}{!}{
\begin{tabular}{c|c|c|c}
Hyperparameter &PT &PFT &VIT \\ \hline
learning rate &1e-3 &2e-6 &4e-6  \\
lr schedule &Cosine &Cosine & Cosine \\
batchsize per GPU &32 &8 &8 \\
GPUs &8$\times$A100 &16$\times$A100 &32$\times$A100 \\
Zero &Zero2 &Zero3 &Zero3-offload \\
Optimizer &AdamW &AdamW &AdamW \\
MLP &Open &Open &Open \\
CLIP &Freeze &Open &Open \\ 
LLM &Freeze &Open &Open \\
MoE blocks &- &- &\checkmark \\
Max Token &2048 &4096 &4096 \\ \hline
\end{tabular}}
\caption{Hyperparameters used in three-stage training on Mistral-7B. PT: Pre-Training stage. PFT: Pre-FineTuning stage. VIT: Visual Instruction tuning stage.} 
\label{tab:app_hyper}
\end{table}
\begin{table}[htb]
\centering
\resizebox{0.5\textwidth}{!}{
\begin{tabular}{c|c|c|c|c}
CuMo &CLIP &MLP &LLM &Total \\ \hline
Mistral-7B &0.30B &0.025B &7.25B  &7.58B \\
$\rightleftharpoons$ Activation Params &0.30B &0.025B &7.25B &7.58B \\
+ Top 2-in-4 MLP-MoE &0.30B &0.10B &7.25B &7.65B \\
$\rightleftharpoons$ Activation Params &0.30B &0.05B &7.25B &7.60B \\
+ Top 2-in-4 CLIP-MoE  &0.91B &0.10B &7.25B &8.26B  \\
$\rightleftharpoons$ Activation Params &0.50B &0.05B &7.25B &7.80B \\ 
$\rightleftharpoons$ Mixtral-8x7B &0.91B &0.10B &46.70B &47.71B \\
$\rightleftharpoons$ Activation Params &0.50B &0.05B &12.90B &13.45B \\ \hline
\end{tabular}}
\caption{Change of model parameters of CuMo. The 7.80B and 13.45B activation parameters corresponds to Act. of CuMo in Table~\ref{tab:m3_final}.} 
\label{tab:app_params}
\end{table}

\section{Experimental Setup Details}
\label{appendix:setup}
Table~\ref{tab:app_hyper} provides an overview of the main hyperparameters used during the three-stage training process. For the final results presented in Table~\ref{tab:m3_final}, the model was trained using 32 $\times$ A100 GPUs with a total batch size of 256 and a learning rate of 4e-6. All ablation studies were conducted with a total batch size of 128 and learning rates of 2e-5 and 2e-6, as detailed in Section~\ref{4.3}.

\begin{figure*}[tb]
\centering
\includegraphics[width=1.0\textwidth]{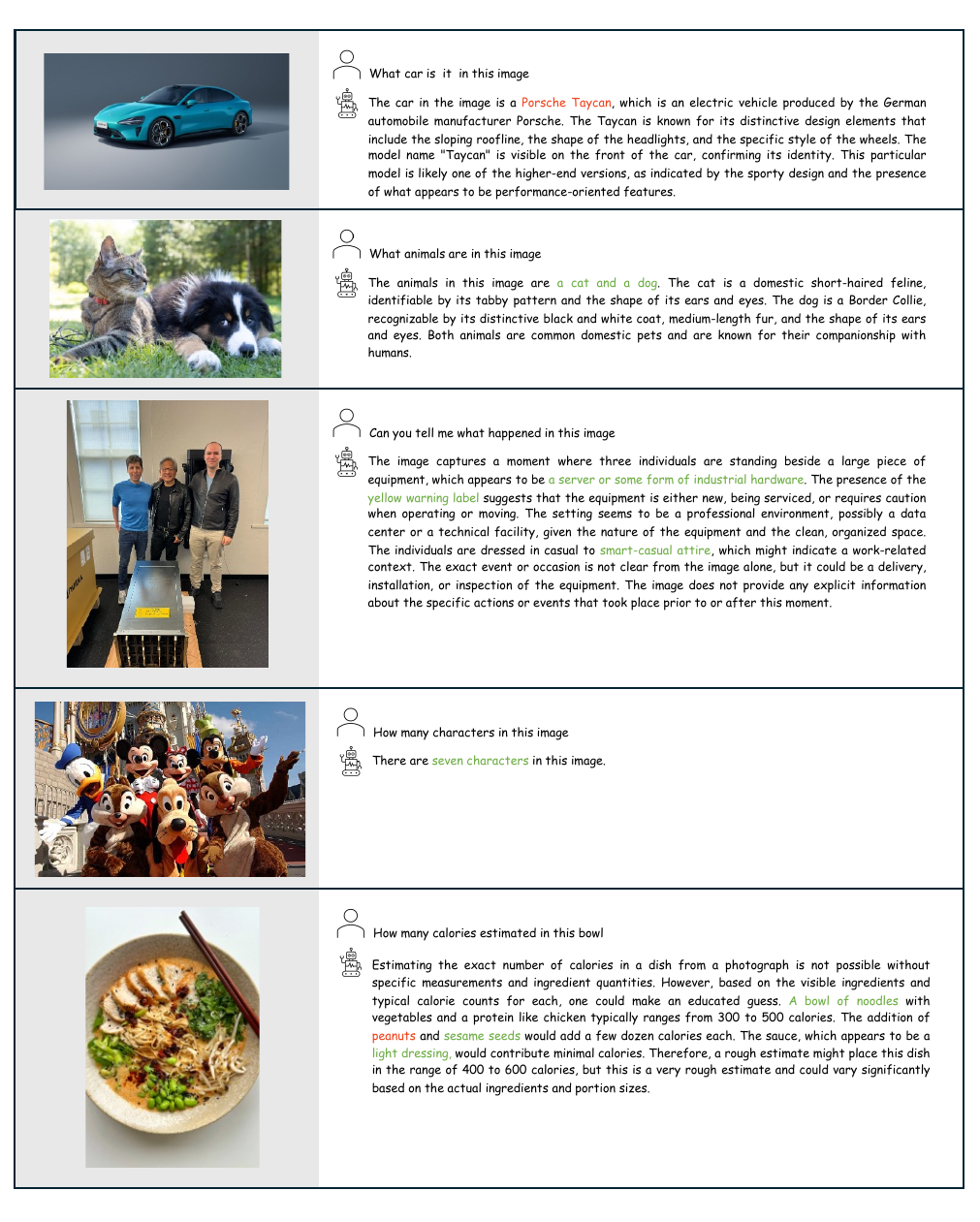}
\caption{\textbf{More dialogues between the user and CuMo.} We highlight the \textcolor{demogreen}{correct} answers and \textcolor{demored}{hallucinations} from the responses of CuMo.}
\label{fig:demo_app}
\end{figure*}

\section{Model Parameters}
We include Table~\ref{tab:app_params} to illustrate the evolution of parameters in the CuMo model throughout its construction process. The LLM constitutes a significant proportion of the total parameters, underscoring the potential for further scaling up the vision encoders to bolster the strength of multimodal LLMs.

\section{More Dialogues}
\label{appendix:dialogues}
We add more dialogues between the questions from user and the response from CuMo in Figure~\ref{fig:demo_app}.
\end{document}